\documentclass[aps,prx,reprint,superscriptaddress,longbibliography,floatfix]{revtex4-2}
\usepackage[T1]{fontenc}
\usepackage{graphicx}
\usepackage{amsmath,amssymb}
\usepackage{xurl}
\usepackage{booktabs}
\usepackage{placeins}
\usepackage[colorlinks=true,urlcolor=blue,citecolor=blue,linkcolor=blue]{hyperref}
\DeclareGraphicsExtensions{.pdf,.png,.eps,.jpg}
\begin{document}

\title{A Hamiltonian-Inspired Local-Operator Ansatz for Slimming Large Language Models}

\author{Ying Lu}
\affiliation{School of Physical Sciences, University of Chinese Academy of Sciences, Beijing 100049, China}
\affiliation{Kavli Institute for Theoretical Sciences, University of Chinese Academy of Sciences, Beijing 100190, China}
\author{Peng-Fei Zhou}
\affiliation{Center for Quantum Physics and Intelligent Sciences, Department of Physics, Capital Normal University, Beijing 100048, China}
\author{Qi-Xuan Fang}
\affiliation{School of Physical Sciences, University of Chinese Academy of Sciences, Beijing 100049, China}
\affiliation{Kavli Institute for Theoretical Sciences, University of Chinese Academy of Sciences, Beijing 100190, China}
\author{Pan Zhang}
\affiliation{Institute of Theoretical Physics, Chinese Academy of Sciences, Beijing 100190, China}
\author{Shi-Ju Ran}
\email{sjran@cnu.edu.cn}
\affiliation{Center for Quantum Physics and Intelligent Sciences, Department of Physics, Capital Normal University, Beijing 100048, China}
\author{Gang Su}
\email{sugang@itp.ac.cn}
\affiliation{Institute of Theoretical Physics, Chinese Academy of Sciences, Beijing 100190, China}
\affiliation{School of Physical Sciences, University of Chinese Academy of Sciences, Beijing 100049, China}
\affiliation{Kavli Institute for Theoretical Sciences, University of Chinese Academy of Sciences, Beijing 100190, China}

\begin{abstract}
Dense linear maps carry much of the parameter and computational burden of modern neural networks, yet their dense form leaves the organization of learned couplings implicit. Quantum many-body physics organizes exponentially large operators by writing a global Hamiltonian as a sum of local terms, \(\hat H=\sum_k\hat h_k\). Whether the same structural principle can carry learned neural maps is unknown. We introduce Tensor Mixture (MixT), which represents a dense map as a natively executable sum of overlapping local tensor operators without imposing an explicit matrix-rank constraint. The local-term count \(N_T\) sets the effective nonlocality and operator complexity, while the number of replaced Transformer blocks \(N_B\) extends this structural coordinate across network depth. Tests on Qwen3-8B and LLaMA2-7B reveal a broad recoverable regime followed by an abrupt, model-specific boundary that is remarkably stable against changes in \(N_T\). Accuracy and output-distribution statistics reorganize together across the boundary; in LLaMA2-7B, the same depth separates two scaling regimes of inter-layer geometry drift. The directly executed structure also reduces parameters, arithmetic, storage, and memory. These results establish the local-sum structure as a viable organizing principle for learned linear maps at billion-parameter scale and expose a sharp boundary in their tolerance to structural simplification.
\end{abstract}

\maketitle


\section{Introduction}

\begin{figure*}[tbp]
	\centering
	\includegraphics[width=0.9\linewidth]{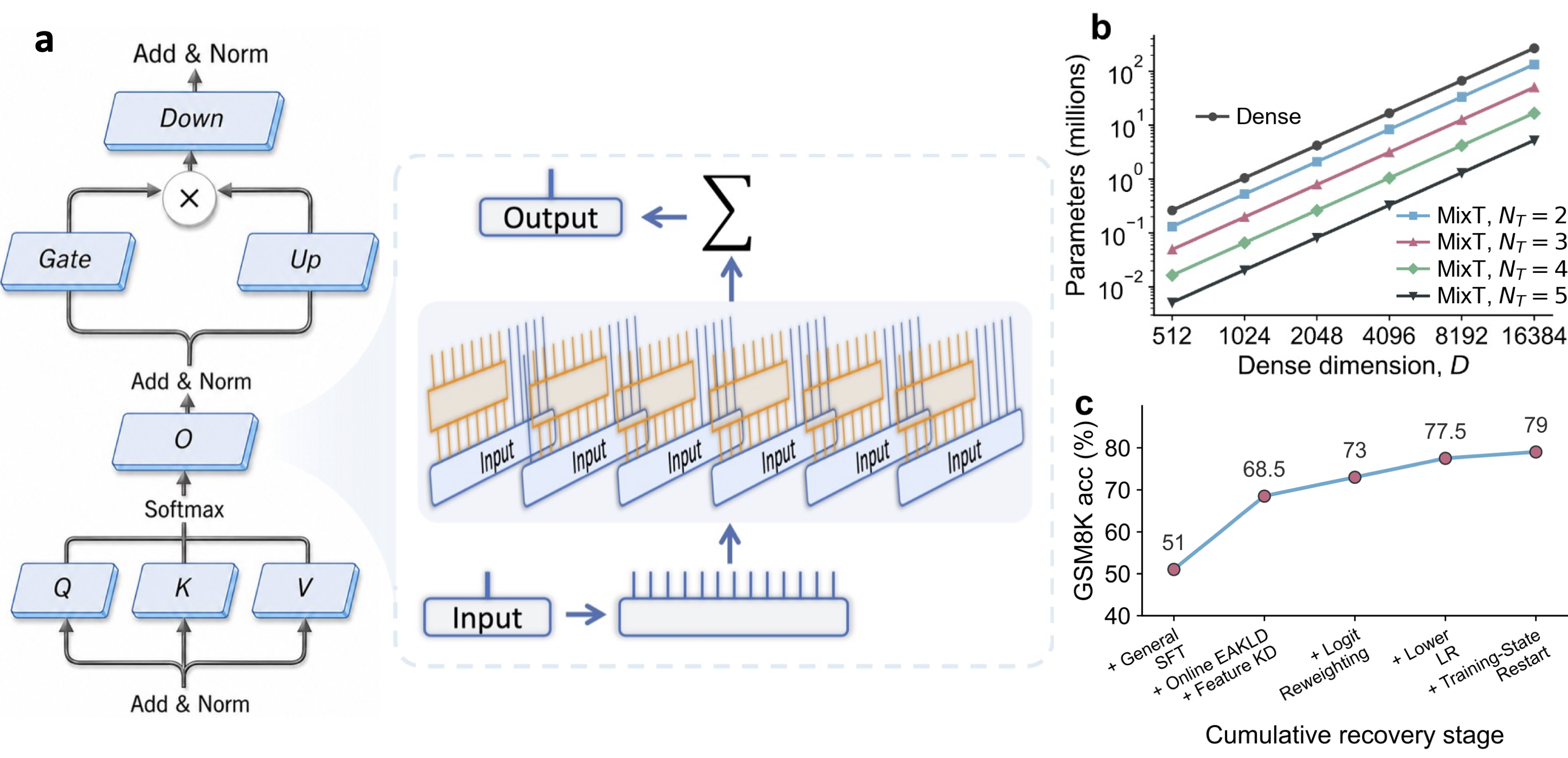}
	\caption{
		MixT transfers the Hamiltonian sum-of-local-terms structure to Transformer maps.
		(a) Dense attention and feed-forward projections are replaced by sums of overlapping local tensor operators while the residual topology is retained.
		(b) Operator-level parameter scaling with matrix dimension \(D\) and local-term count \(N_T\).
		(c) Successive recovery stages improve GSM8K performance at fixed hybrid MixT structure.}
	\label{fig:mixt_overview}
\end{figure*}

The rapid development of large language models (LLMs) has been accompanied by steep growth in parameter count, storage, memory demand, and computation~\cite{NEURIPS2022_c1e2faff,Xiao_2025}. Dense linear projections carry a substantial part of this burden, from the query, key, value, and output maps of self-attention to the gate, up, and down maps of feed-forward blocks~\cite{NIPS2017_3f5ee243,Qiu2024BTT}. Their dense parameterization also conceals structural organization: it states that every output may depend on every input, but not which couplings are indispensable or how the map can be simplified in a controlled manner. Quantization, pruning, low-rank replacement, and layer removal expose different forms of redundancy~\cite{MLSYS2024_42a452cb,frantar2023gptqaccurateposttrainingquantization,pmlr-v202-xiao23c,pmlr-v202-frantar23a,NEURIPS2023_44956951,xia2024shearedllamaacceleratinglanguage,hu2021loralowrankadaptationlarge,Ding_2023,chen2025streamliningredundantlayerscompress,gromov2025}. A more basic question is whether the linear map itself admits an executable organization whose complexity can be varied as a structural coordinate.

Quantum many-body physics has long faced this operator problem in exponentially large Hilbert spaces. Mean-field descriptions reduce the many-body problem to self-consistent one-body structure, while tensor networks exploit low-rank factorizations of states and operators whose bond dimensions control correlation capacity and cost~\cite{Slater1963HartreeFock,Schollwock2011,ORUS2014117,Pirvu2010,TensorDecompositionsandApplications}. Another fundamental operator-level principle is locality: a lattice Hamiltonian is written as a sum of local terms,
\(
\hat H=\sum_k\hat h_k,
\)
with each \(\hat h_k\) acting only on a bounded subspace~\cite{BravyiHastingsVerstraete2006,GongGuaitaCirac2023}. Factorization controls the complexity carried across partitions; this local-sum structure controls how a global operator is assembled from supported pieces. Both replace an unconstrained exponential object by an explicit structural construction.

The former low-rank route has begun to enter artificial intelligence through tensor-train, Tucker, and matrix-product-operator (MPO) layers for neural-network compression, language modeling, and adaptation~\cite{NIPS2015_6855456e,NEURIPS2019_dc960c46,Gao2020MPOCompression,Liu2021MPOFineTuning,yang-etal-2024-loretta}. Low-rank, Kronecker, butterfly, Monarch, and block-tensor constructions provide further structured parameterizations~\cite{ZhouWu2015,Tahaei2021KroneckerBERT,Dao2019Butterfly,Dao2022Monarch,Qiu2024BTT,Compacter2021,ProcrustesGPT2025}, and tensor networks can themselves serve as learning models~\cite{NIPS2016_5314b967,ran2023tensor,qing2025compressing}. These approaches primarily control matrix rank, tensor rank, bond dimension, or global factor structure. By contrast, the local-sum structure---a learned map represented and executed as a sum of identity-extended local operators---has not been established as a neural linear-map architecture at LLM scale. This leaves open whether local operators can carry learned global maps and where their tolerance breaks down when the ansatz is imposed through network depth.

Here we introduce Tensor Mixture (MixT), formulate this local-sum structure as a neural linear-map ansatz, and realize it as a natively executable operator for billion-parameter LLMs. After tensorizing the feature coordinates, MixT represents a global linear map as a normalized sum of overlapping local tensor operators. The local-term count \(N_T\) sets their effective nonlocality and operator-level complexity, while the number of replaced Transformer blocks \(N_B\) controls how deeply the ansatz is imposed. Experiments on Qwen3-8B~\cite{yang2025qwen3technicalreport} and LLaMA2-7B~\cite{touvron2023llama2openfoundation} show that the local-operator ansatz carries learned maps across a broad range of replacement depths before reaching an abrupt, model-specific boundary. The boundary remains nearly unchanged across a wide range of \(N_T\), while accuracy and output distributions reorganize in both models and inter-layer geometry crosses between two scaling regimes in LLaMA2-7B. Because the local terms remain directly executable, the same structure produces systematic reductions in parameters, arithmetic, storage, and memory. MixT thus establishes the local-sum structure as a neural operator principle at billion-parameter scale and turns progressive model slimming into a controlled probe of structural reorganization.
  
\section{Tensor Mixture as a local-operator ansatz}

\subsection{Construction}

MixT replaces a dense linear operator \(\hat F:\mathbb{R}^{D_{\mathrm{in}}}\to\mathbb{R}^{D_{\mathrm{out}}}\) by a sum of local tensor operators [Fig.~\ref{fig:mixt_overview}(a)]. In the selected Transformer blocks~\cite{NIPS2017_3f5ee243}, we replace the \(Q\), \(K\), \(V\) and \(O\) projections in self-attention and the gate, up and down maps in the feed-forward module. We first consider tensorizable dimensions \(D_{\mathrm{in}}=d^n\) and \(D_{\mathrm{out}}=d^m\), where \(d\) is the local site dimension and \(d=2\) throughout this study. Tensorization maps feature indices to an ordered set of synthetic sites, each associated with a local \(d\)-dimensional space. These sites define the support structure of the ansatz rather than a physical lattice. A hybrid extension for non-dyadic maps is described in the Supplemental Material~\cite{SupplementalMaterial}, Sec.~S5.

For a chosen number of local terms \(N_T\le \min(n,m)\), the MixT operator is
\begin{equation}
\begin{aligned}
\hat F_{\mathrm{MixT}}
&=\frac{1}{N_T}\sum_{k=1}^{N_T}\hat L_k,\\
\hat L_k
&=\hat I_d^{\otimes(k-1)}\otimes\hat T^{[k]}
\otimes\hat I_d^{\otimes(N_T-k)} .
\end{aligned}
\label{eq:mixt_operator}
\end{equation}
Here \(\hat T^{[k]}\) acts on the \(k\)th contiguous input and output supports, while identity operators act on the remaining \(N_T-1\) synthetic sites. Successive supports are distinct and generally overlap. Thus \(N_T\) controls both the number and the support range of the local terms. The factor \(1/N_T\) fixes the initialization scale and can be absorbed into the local operators. Exact support assignments, index contractions, and representative choices of \(N_T\) are given in the Supplemental Material~\cite{SupplementalMaterial}, Sec.~S1.

Eq.~(\ref{eq:mixt_operator}) transfers the sum-of-local-terms organization of a Hamiltonian to a learned map. In \(\hat H=\sum_k\hat h_k\), each \(\hat h_k\) acts only on a bounded set of sites but contributes to a global operator; in MixT, \(\hat L_k\) plays the same structural role on synthetic sites. The correspondence concerns operator organization, not dynamics: the local maps need not be square or Hermitian. In the square case, \(N_T=1\) recovers a general dense map, whereas increasing \(N_T\) produces more numerous terms of progressively shorter support.

\subsection{Parameter and arithmetic scaling}

The local-term count and support widths jointly determine the storage and arithmetic complexity of the ansatz [Fig.~\ref{fig:mixt_overview}(b)]. Each local tensor spans
\(
w=n-N_T+1
\)
input sites and
\(
w'=m-N_T+1
\)
output sites. Excluding biases, the parameter counts are
\begin{equation}
\begin{aligned}
\#\mathrm{par}(\hat F_{\mathrm{dense}})
&=D_{\mathrm{in}}D_{\mathrm{out}}=d^{m+n},\\
\#\mathrm{par}(\hat F_{\mathrm{MixT}})
&=N_Td^{w+w'}=N_T d^{m+n+2-2N_T}.
\end{aligned}
\label{eq:mixt_param_count}
\end{equation}
The retained parameter fraction is therefore
\begin{equation}
r_{\mathrm{par}}
\equiv
\frac{\#\mathrm{par}(\hat F_{\mathrm{MixT}})}
{\#\mathrm{par}(\hat F_{\mathrm{dense}})}
=\frac{N_T}{d^{2N_T-2}} .
\label{eq:mixt_param_ratio}
\end{equation}
Its numerator grows only linearly with \(N_T\), whereas its denominator grows exponentially. The retained parameter fraction therefore decreases exponentially with the number of local terms.

The same local-support restriction also reduces arithmetic. Each local tensor is applied across the \(d^{N_T-1}\) configurations of the identity-passed sites, giving the dominant multiply--accumulate (MAC) counts
\begin{equation}
\begin{aligned}
\#\mathrm{MAC}(\hat F_{\mathrm{dense}})&=d^{m+n},\\
\#\mathrm{MAC}(\hat F_{\mathrm{MixT}})
&=N_Td^{w+w'}d^{N_T-1}
=N_Td^{m+n+1-N_T}.
\end{aligned}
\label{eq:mixt_mac_count}
\end{equation}
\begin{equation}
r_{\mathrm{MAC}}
\equiv
\frac{\#\mathrm{MAC}(\hat F_{\mathrm{MixT}})}
{\#\mathrm{MAC}(\hat F_{\mathrm{dense}})}
=\frac{N_T}{d^{N_T-1}} .
\label{eq:mixt_mac_ratio}
\end{equation}
For \(d=2\) and \(N_T=4\), a \((4096,4096)\) map is reduced from \(16.78\) million to \(1.05\) million parameters and from \(16.78\) million to \(8.39\) million MACs, corresponding to \(r_{\mathrm{par}}=6.25\%\) and \(r_{\mathrm{MAC}}=50\%\).
The arithmetic ratio has the analogous linear-over-exponential form, with a different exponent because each local tensor is reused across the identity-passed sites. These counts exclude the lower-order additions required to sum the \(N_T\) outputs. At whole-model scale, unreplaced blocks and nonlinear operations reduce the arithmetic gain relative to the operator-level ratio.

MixT derives compactness from a restriction on operator organization rather than from an explicit truncation of matrix rank. The resulting local-sum operator nevertheless admits an exact representation within the established matrix-product-operator (MPO) framework. A standard finite-state construction encodes each supported local term as a path through a single MPO, and the sum over paths reproduces \(\hat F_{\mathrm{MixT}}\) exactly~\cite{CrosswhiteBacon2008,Pirvu2010,Hubig2017}. This correspondence establishes the tensor-network representability of MixT; it does not define either its native parameters or its forward contraction. The construction and the corresponding bond-dimension bounds are given in the Supplemental Material~\cite{SupplementalMaterial}, Sec.~S1.

The local-sum ansatz imposes no explicit matrix-rank bottleneck. Up to input and output permutations,
\(\hat L_k=\hat T^{[k]}\otimes\hat I_{d^{N_T-1}}\), and hence
\(\operatorname{rank}(\hat L_k)
=d^{N_T-1}\operatorname{rank}(\hat T^{[k]})\).
A maximal-rank local map therefore yields
\(\operatorname{rank}(\hat L_k)
=\min(D_{\mathrm{in}},D_{\mathrm{out}})\), while their sum is generically full rank. MixT thus achieves the parameter reduction in Eq.~(\ref{eq:mixt_param_ratio}) by constraining local operator organization rather than matrix rank. Moreover, its \(N_T\) local contractions constitute the native forward map: they admit parallel evaluation and can be summed without reconstructing a dense matrix. This combination of compact parametrization, potentially full matrix rank, and direct local-sum execution distinguishes MixT from stand-alone low-rank replacements and tensorized layers that rely on serial chain contractions or dense reconstruction~\cite{NEURIPS2019_dc960c46,Gao2020MPOCompression,Liu2021MPOFineTuning,javanmard2026compressingtransformerlanguagemodels}.

\begin{figure*}[!t]
	\centering
	\includegraphics[width=0.9\linewidth]{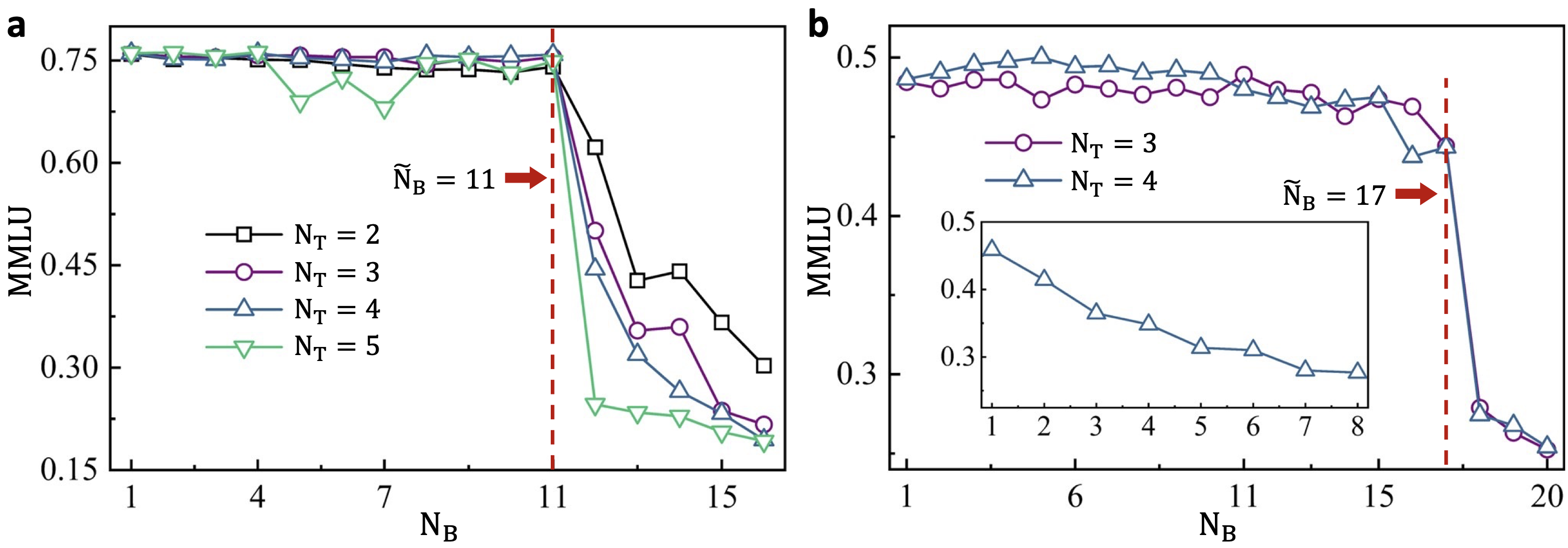}
	\caption{An abrupt structural boundary under progressive local-operator replacement. (a) Qwen3-8B MMLU accuracy under back-to-front replacement for different local-term counts \(N_T\); the dashed line marks \(\tilde N_B=11\). (b) LLaMA2-7B sweep with \(\tilde N_B=17\); the inset shows front-to-back replacement.}
	\label{fig:transition_scan}
\end{figure*}

\section{Compression and diagnostic protocol}

\subsection{Structural replacement and recovery}

All main-text experiments used \(d=2\) and replaced the \(Q\), \(K\), \(V\), \(O\), gate, up, and down maps in each selected Transformer block. Token embeddings, the output head, and normalization layers retained their dense structure. For LLaMA2-7B, the replaced feed-forward maps operate in a padded \(11008\to16384\) intermediate space; the reported parameter and arithmetic counts include this overhead. Implementation details for padded and non-dyadic maps are given in the Supplemental Material~\cite{SupplementalMaterial}, Secs.~S3 and S5.

The replacement depth \(N_B\) counts blocks from the output side toward the input side; a front-to-back sweep was performed for comparison. Each replacement is initialized by matching its dense weight matrix. For a dense matrix \(W\) and the corresponding MixT map \(W_{\mathrm{MixT}}(\{T^{[k]}\})\), the local tensors minimize
\begin{equation}
\min_{\{T^{[k]}\}}\mathcal{L}_{\mathrm{match}},
\qquad
\mathcal{L}_{\mathrm{match}}
=
\left\|
W-W_{\mathrm{MixT}}(\{T^{[k]}\})
\right\|_F^2 .
\label{eq:weight_matching}
\end{equation}
The resulting checkpoint is then recovered under a protocol held fixed within each model family. MixT parameters in replaced blocks are trainable, whereas unreplaced dense weights, token embeddings, and the output head remain fixed; normalization parameters in replaced LLaMA2 blocks are also updated. Optimization settings, recovery data, and evaluation protocols are specified in the Supplemental Material~\cite{SupplementalMaterial}, Sec.~S2.

\subsection{Operational boundary and output diagnostics}

General reasoning performance was evaluated on MMLU~\cite{hendrycks2021measuring} with the Language Model Evaluation Harness~\cite{biderman2024lessonstrenchesreproducibleevaluation}. Let \(A_P(N_B)\) denote the recovered accuracy under protocol \(P\), \(A_0\) the dense accuracy, and \(\mathcal{S}\) the scanned replacement depths. We define the operational boundary as
\begin{equation}
\begin{aligned}
\tilde N_B(P;\delta)
&=\max_{\substack{N_B\in\mathcal{S}\\A_P(N_B)\ge A_0-\delta}}N_B,\\
\delta&=0.03,
\end{aligned}
\label{eq:operational_boundary}
\end{equation}
with subsequent scanned depths required to remain below the criterion. Thus \(\tilde N_B\) is the last recoverable point under the fixed protocol.

For each MMLU question, final-step logits restricted to the answer tokens A--D define a normalized distribution \(\{p_j\}\). Its mean normalized entropy is
\begin{equation}
\mathrm{OE}=\frac{\mathbb{E}[H_4]}{\log 4},\qquad
H_4(p)=-\sum_{j\in\{A,B,C,D\}}p_j\log p_j .
\label{eq:oe}
\end{equation}
To measure dataset-level answer usage, let \(\hat y\) be the predicted label and \(f_j=\Pr(\hat y=j)\) its empirical frequency. The normalized prediction entropy is
\begin{equation}
\mathrm{PE}=\frac{H_{\mathrm{pred}}}{\log 4},\qquad
H_{\mathrm{pred}}=-\sum_{j\in\{A,B,C,D\}}f_j\log f_j .
\label{eq:pe}
\end{equation}
OE therefore measures per-question uncertainty, whereas PE measures answer diversity across the evaluation set.

\subsection{Inter-layer geometry}

We track representation reorganization through inter-layer similarity at the answer-decision position. For a fixed prompt set \(\mathcal{Q}\), let \(h^{(N_B)}_{\ell}(q)\) be the hidden state at layer \(\ell\) for prompt \(q\). The similarity map is
\[
S_{N_B}(\ell,\ell')
=\frac{1}{|\mathcal{Q}|}
\sum_{q\in\mathcal{Q}}
\cos\!\left[h^{(N_B)}_{\ell}(q),h^{(N_B)}_{\ell'}(q)\right],
\]
and its drift from the dense reference \(S_0\) is
\begin{equation}
\Delta S_{N_B}(\ell,\ell')
=\left|S_{N_B}(\ell,\ell')-S_0(\ell,\ell')\right| .
\label{eq:deltas}
\end{equation}
For a set of layer pairs \(\mathcal{P}\), the aggregate drift is
\begin{equation}
\overline{\Delta S}_{N_B}(\mathcal{P})
=\frac{1}{|\mathcal{P}|}
\sum_{(\ell,\ell')\in\mathcal{P}}
\Delta S_{N_B}(\ell,\ell') .
\label{eq:geometry_average}
\end{equation}
The prompt set, answer position, and layer-pair selections are specified in the Supplemental Material~\cite{SupplementalMaterial}, Sec.~S4.

\FloatBarrier
\section{Results}

\begin{figure*}[!t]
	\centering
	\includegraphics[width=0.90\linewidth]{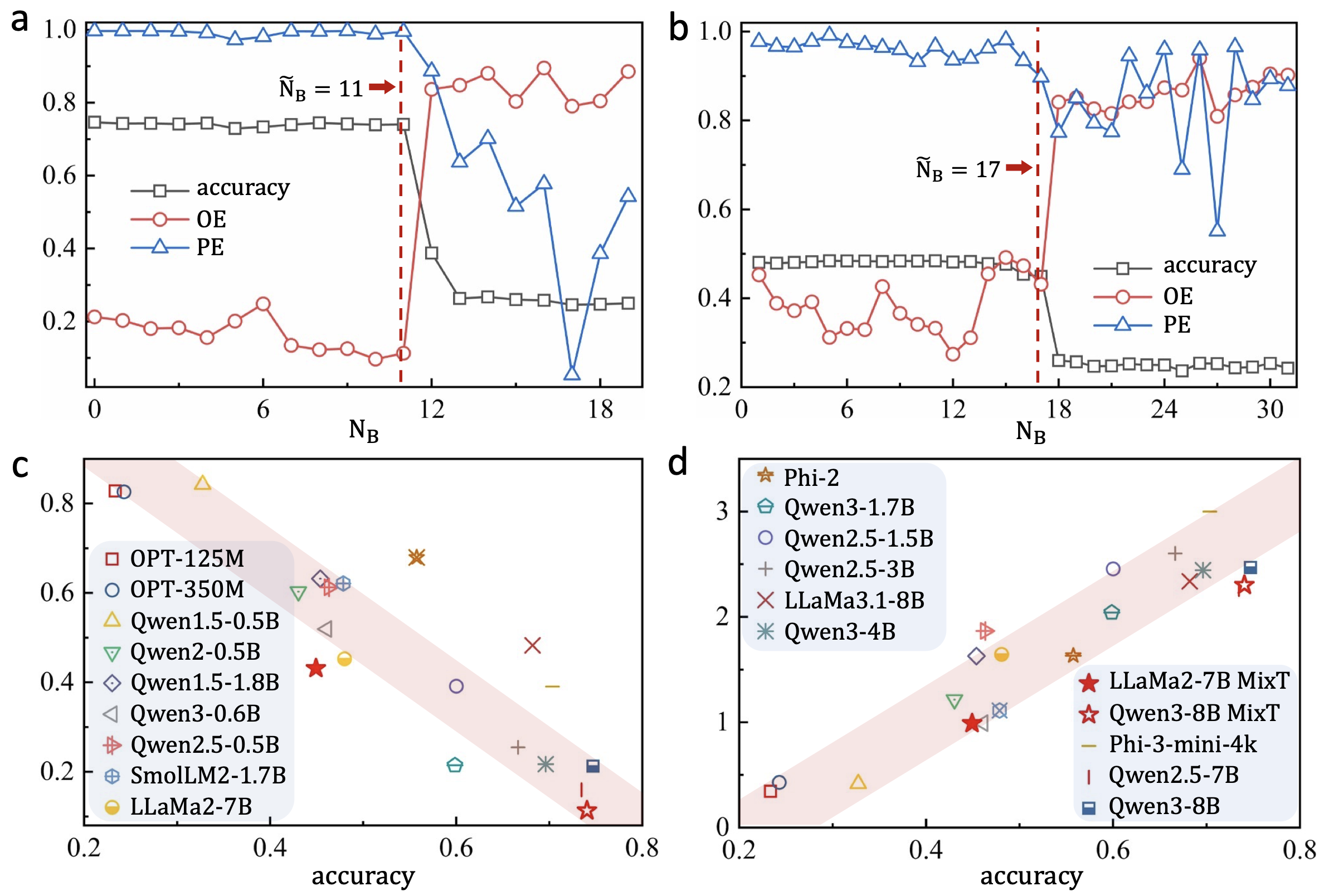}
	\caption{Output distributions reorganize across the structural boundary. (a),(b) MMLU accuracy, output entropy (OE), and prediction entropy (PE) across the Qwen3-8B and LLaMA2-7B sweeps; dashed lines mark their operational boundaries. (c) OE versus MMLU accuracy for dense and MixT checkpoints. (d) Corresponding relation for transformed prediction entropy, \(-\log_{10}(1-\mathrm{PE})\). Shaded bands show empirical linear trends.}
	\label{fig:output_reorganization}
\end{figure*}

\subsection{Progressive local-operator replacement reveals an abrupt boundary}

Progressively imposing the local-operator ansatz from the output side reveals two sharply separated regimes. Under the fixed recovery protocol, we define \(\tilde N_B\) as the last scanned replacement depth whose recovered MMLU accuracy remains within three percentage points of the dense model [Eq.~(\ref{eq:operational_boundary})]. Both LLMs retain a broad recoverable regime up to \(\tilde N_B\), followed by an abrupt transition-like loss of capability (Fig.~\ref{fig:transition_scan}).

For Qwen3-8B, the recoverability boundary occurs at \(\tilde{N}_B=11\), whereas for LLaMA2-7B it occurs at \(\tilde{N}_B=17\). The corresponding dense baselines are \(76.89\%\) for Qwen3-8B and \(46.38\%\) for LLaMA2-7B on MMLU. At the operational boundary, Qwen3-8B remains above \(76.0\%\), and the \(N_T=4\) LLaMA2-7B boundary checkpoint scores \(43.56\%\), so both satisfy the three-percentage-point preservation criterion. Beyond these boundaries, replacing only one or two additional blocks is sufficient to induce large degradation, in some cases reducing accuracy close to the four-way random-guessing level of MMLU. The loss of recoverability is therefore abrupt rather than gradual.

\begin{table}[!t]
	\caption{Resource gains of LLaMA2-7B with increasing local-term count. Both MixT models use \(N_B=17\); \(N_T=4\) is the operational-boundary checkpoint and \(N_T=5\) is evaluated at the same replacement depth. Each change is measured relative to the dense model. FLOPs are analytic estimates, memory is measured, and storage is estimated in GiB. The int8 and int4 entries are storage estimates rather than quantized-accuracy measurements. The measurement protocol is given in the Supplemental Material~\cite{SupplementalMaterial}, Sec.~S3.}
	\label{tab:resource_llama2}
	\centering
	\footnotesize
	\renewcommand{\arraystretch}{1.10}
	\setlength{\tabcolsep}{2.5pt}
	\begin{tabular*}{\columnwidth}{@{}l@{\extracolsep{\fill}}rrrrr@{}}
		\toprule
		\textbf{Metric} & \textbf{Dense} & \multicolumn{2}{c}{\textbf{MixT} \((N_T=4)\)} & \multicolumn{2}{c}{\textbf{MixT} \((N_T=5)\)} \\
		& & \textbf{Value} & \textbf{Change} & \textbf{Value} & \textbf{Change} \\
		\midrule
		Parameters (B) & 6.74 & 3.58 & $46.8\%$ & 3.39 & $49.7\%$ \\
		\midrule
		\multicolumn{6}{@{}l}{FLOPs} \\
		Inference (GFLOPs) & 847.85 & 699.54 & $17.5\%$ & 590.02 & $30.4\%$ \\
		Training (TFLOPs) & 60.32 & 46.08 & $23.6\%$ & 35.57 & $41.0\%$ \\
		\midrule
		\multicolumn{6}{@{}l}{Device memory (GiB)} \\
		Peak inference & 12.61 & 6.79 & $46.2\%$ & 6.42 & $49.1\%$ \\
		Peak training & 38.25 & 9.29 & $75.7\%$ & 8.90 & $76.7\%$ \\
		\midrule
		\multicolumn{6}{@{}l}{Deploy storage (GiB)} \\
		bf16 & 12.55 & 6.67 & $46.8\%$ & 6.31 & $49.7\%$ \\
		int8 & 6.37 & 3.39 & $46.8\%$ & 3.20 & $49.8\%$ \\
		int4 & 3.24 & 1.72 & $46.8\%$ & 1.63 & $49.7\%$ \\
		\bottomrule
	\end{tabular*}
\end{table}

The most striking feature is the stability of the boundary against the microscopic strength of the structural simplification. For Qwen3-8B, \(\tilde{N}_B=11\) throughout \(N_T=2\)--5, even though the parameter reduction of each replaced map increases to above 98\% at \(N_T=5\). The boundary is therefore not set simply by the parameter count or matrix-level capacity of an individual replacement. Its invariance identifies a network-level scale in the back-to-front replacement path. At the largest \(N_T\), the pre-boundary plateau becomes more irregular, but its endpoint does not move.

The inset of Fig.~\ref{fig:transition_scan}(b) shows that compression behavior is strongly direction dependent. When the same procedure is applied from earlier to later layers, performance deteriorates immediately and monotonically, without an extended stable regime or a comparable recoverability boundary. This asymmetry indicates that tolerance to structural simplification is not uniform across depth: output-side blocks are more replaceable, whereas input-side blocks are substantially more sensitive, consistent with previous reports of depth-dependent functional specialization in LLMs~\cite{chen2025streamliningredundantlayerscompress,gromov2025,zhang-etal-2024-investigating,men-etal-2025-shortgpt}.

\begin{figure*}[!t]
	\centering
	\includegraphics[width=0.95\textwidth]{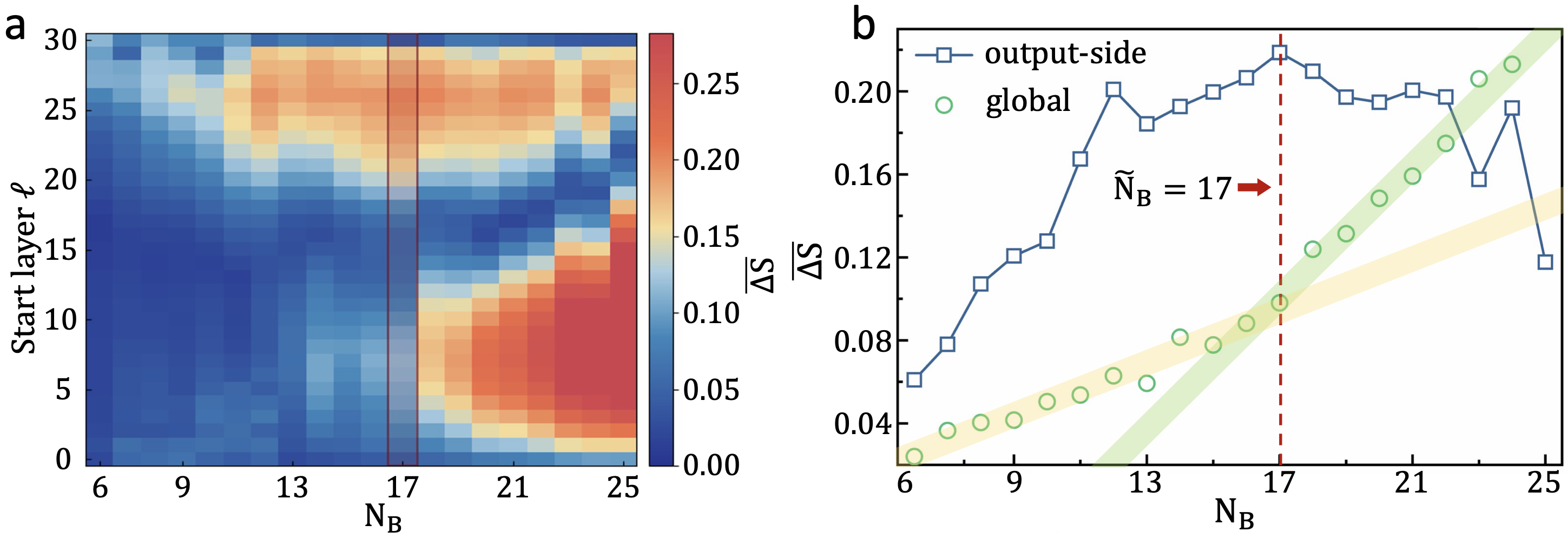}
	\caption{Geometry drift crosses over at the recoverability boundary. (a) LLaMA2-7B drift landscape for the \(N_T=4\) back-to-front sweep; the red band marks \(\tilde N_B=17\). (b) Output-side and global drift averages. Guide bands show slopes of approximately 0.006 and 0.016 per replaced block. Sampling and aggregation are specified in the Supplemental Material~\cite{SupplementalMaterial}, Sec.~S4.}
	\label{fig:interlayer_structure}
\end{figure*}

The reorganization is selective across capabilities rather than uniform (Supplemental Material~\cite{SupplementalMaterial}, Sec.~S2). For LLaMA2-7B, MMLU and GSM8K remain close to their dense references at \(\tilde N_B=17\) but fall sharply at \(N_B=19\). Qwen3-8B shows the same abrupt losses between \(\tilde N_B=11\) and \(N_B=13\), while other representative tasks decline more moderately. Multiple capability probes therefore change near the structural boundary, but with task-dependent amplitudes.

At fixed hybrid MixT structure in Qwen3-4B-Instruct, successive recovery stages recover substantial additional GSM8K capability [Fig.~\ref{fig:mixt_overview}(c)]. The trajectory exposes optimization headroom within the ansatz and complements the structural boundary scan.

\subsection{Native execution yields systematic resource reductions}

The local-operator representation remains present in the executable model and produces systematic resource reductions at the recoverability boundary (Table~\ref{tab:resource_llama2}). At \(N_T=4\), effective parameters decrease by \(46.8\%\), inference FLOPs by \(17.5\%\), and peak inference memory and bf16 storage by about \(46\%\). At the same replacement depth, increasing the local-term count to \(N_T=5\) raises these reductions to \(49.7\%\), \(30.4\%\), \(49.1\%\), and \(49.7\%\), respectively, while remaining within the MMLU-preserving regime. This monotonic response establishes \(N_T\) as a direct control over the cost of the local-operator ansatz; unchanged model components account for the smaller whole-model gains relative to the operator-level scaling.

\subsection{Output distributions reorganize across the structural boundary}

The capability boundary coincides with a reorganization of the model output, not only a change in discrete accuracy. For both Qwen3-8B and LLaMA2-7B, crossing \(\tilde N_B\) produces an abrupt accuracy drop together with sharp changes in two continuous statistics of the MMLU answer distribution (Fig.~\ref{fig:output_reorganization}). Output entropy (OE) and prediction entropy (PE) therefore resolve complementary aspects of the same structural sweep beyond the benchmark score alone~\cite{Wei2022EmergentAbilities,Schaeffer2023Mirage}.

Figure~\ref{fig:output_reorganization}(a),(b) first shows a clear change in output entropy. Before \(\tilde N_B\), accuracy remains relatively stable and OE, defined in Eq.~\ref{eq:oe}, stays low, indicating that the model usually concentrates probability mass on a small subset of answer options for each question. Once \(\tilde N_B\) is crossed, OE rises abruptly while accuracy falls. The recoverability boundary therefore marks a sudden loss of per-question decisiveness, rather than a gradual weakening of performance alone.

Prediction entropy behaves differently. OE is defined from the full four-way answer distribution on each MMLU question, whereas PE, defined in Eq.~\ref{eq:pe}, is computed from the distribution of final predicted labels over the full evaluation set. In both sweeps, PE is high before the boundary and drops after it. This means that the post-boundary model becomes not only more uncertain on individual questions, but also more concentrated in its final label usage across the dataset.

To place these compressed checkpoints in a broader context, Fig.~\ref{fig:output_reorganization}(c),(d) compares them with a wider set of LLMs spanning different scales. Across this model set, OE decreases approximately linearly with MMLU accuracy, whereas the transformed prediction entropy, \(-\log_{10}(1-\mathrm{PE})\), increases approximately linearly. Within the observed accuracy range, and using MMLU accuracy as a fraction, the empirical relations are
\[
\begin{aligned}
\mathrm{OE} &\approx 1.20 - 1.30\,\mathrm{Acc},\\
-\log_{10}(1-\mathrm{PE}) &\approx -0.90 + 4.87\,\mathrm{Acc}.
\end{aligned}
\]
Across scales, MixT checkpoints follow the same empirical trends as the dense models rather than departing from them: compression moves a model along the accuracy--entropy relations instead of away from them. The output statistics observed near the recoverability boundary are therefore not isolated anomalies of a single compression sweep, but part of a broader relation between answer-distribution structure and task performance.

\subsection{Geometry drift crosses over at the recoverability boundary in LLaMA2-7B}

Figure~\ref{fig:interlayer_structure} shows how inter-layer geometry changes across the LLaMA2-7B compression sweep. For each checkpoint, we compare the inter-layer similarity map with that of the dense reference and measure their absolute deviation by \(\Delta S_{N_B}(\ell,\ell')\), defined in Eq.~\ref{eq:deltas}. Larger values indicate larger departures of the corresponding layer pair from the dense model.

To visualize the full compression path, we average \(\Delta S_{N_B}(\ell,\ell')\) over all later layers \(\ell'>\ell\) for each start layer \(\ell\). The resulting landscape shows weak drift at early compression depths and increasingly structured drift near \(\tilde N_B=17\) [Fig.~\ref{fig:interlayer_structure}(a)]. Beyond the boundary, high-drift regions extend across a broader range of start layers.

To quantify these changes, Figure~\ref{fig:interlayer_structure}(b) compares two summary statistics: mean drift over layer pairs near the output end and mean drift over all valid layer pairs. The output-side average rises with compression and reaches a high value around the recoverability boundary. By contrast, the global average is well described by two empirical trend bands: a pre-boundary regime with a scaling coefficient of about 0.006 per compressed block, and a post-boundary regime with a larger coefficient of about 0.016 per compressed block. These two regimes share a linear scaling form but differ in scaling coefficient. Their intersection occurs near \(N_B=\tilde N_B\), identifying the operational boundary as a crossover in the scaling of global geometry drift.

\section{Discussion and conclusions}

The central result is that the local-sum structure of a many-body Hamiltonian can also carry learned neural linear maps at billion-parameter scale. MixT realizes this principle as an explicit forward operator: overlapping local terms are extended by identities, evaluated directly, and summed into the global map. The broad recoverable regimes in Qwen3-8B and LLaMA2-7B establish the representational viability of this ansatz, while the reductions in parameters, arithmetic, storage, and memory establish its operational consequence. The physics connection is therefore structural rather than metaphorical: the same local-term construction defines both the object being tested and the executable model.

Progressive replacement reveals that tolerance to the local-sum ansatz is organized primarily across network depth. In Qwen3-8B, changing \(N_T\) from 2 to 5 strongly alters the support and parameter count of every local map but leaves the boundary fixed at \(\tilde N_B=11\). By contrast, changing \(N_B\) by only one or two blocks beyond the boundary causes a large capability loss, and reversing the replacement direction removes the extended recoverable regime. This separation between sensitivity to structural depth and robustness to local-term complexity identifies the boundary with block-level functional organization rather than a smooth exhaustion of single-map capacity.

The boundary is also visible in several descriptions of model function. Knowledge and mathematical reasoning probes show abrupt losses, other tasks degrade more moderately, and the MMLU answer distribution changes in both per-question uncertainty and dataset-level label usage. In LLaMA2-7B, the same replacement depth separates two scaling regimes of global geometry drift. We therefore distinguish the abrupt, transition-like capability threshold from the crossover in geometry scaling: the former marks loss of recoverability, while the latter describes a change in the rate of internal reorganization. Related transition-like behavior has been observed under neural-network pruning~\cite{pesce2026phasetransitionsneuralnetworks,Pan2026PruningPhases}; MixT exposes it along a Hamiltonian-form structural coordinate and connects it to output and representation geometry. Establishing thermodynamic criticality would require finite-size scaling, an independently defined order parameter, and data collapse, which are separate from the transition-like phenomenology established here.

The relative boundary depths, \(11/36\) for Qwen3-8B and \(17/32\) for LLaMA2-7B, further motivate \(\tilde N_B/L\) as a probe of structural redundancy. A shallower boundary may accompany greater capability density~\cite{Xiao_2025}, a hypothesis that can be tested across model families and recovery budgets. The fixed-structure trajectory in Fig.~\ref{fig:mixt_overview}(c) shows that further capability can be recovered without altering the local-operator ansatz. Hardware-aware kernels, the non-dyadic hybrid construction demonstrated on Qwen3-4B-Instruct (Supplemental Material~\cite{SupplementalMaterial}, Sec.~S5), and independent quantization provide complementary routes to extend the executable gains. A routed variant could further combine local-operator structure with conditional computation~\cite{Shazeer2017MoE,Fedus2022Switch}.

\begin{acknowledgments}
This work was supported in part by the Innovation Program for Quantum Science and Technology (Grant No. 2024ZD0300500), the Strategic Priority Research Program of the Chinese Academy of Sciences (Grant No. XDB1270000), NSFC (Grant No.12534009), and CAS. The numerical simulations were partially performed on the robotic AI-Scientist platform of the Chinese Academy of Sciences.
\end{acknowledgments}

\section*{Data and Code Availability}
The benchmark and recovery datasets are publicly available from their original providers; their composition and evaluation protocols are specified in the Supplemental Material. Before submission, the MixT implementation, structural-replacement and weight-matching code, recovery and evaluation scripts, processed outputs, and numerical data underlying the figures and tables will be deposited in a persistent public repository [URL].

\bibliography{name}

\end{document}